\documentclass[letterpaper, 10 pt, conference]{ieeeconf}  

\IEEEoverridecommandlockouts                              

\usepackage{graphicx}
\usepackage{array}
\usepackage{amsmath} 
\usepackage{cite}
\usepackage{color}
\usepackage{booktabs}
\usepackage{makecell}
\usepackage{tabularx}

\newcommand{\vavg}{\ensuremath{v_{\text{avg}}}}
\newcommand{\vmin}{\ensuremath{v_{\text{min}}}}
\newcommand{\vmax}{\ensuremath{v_{\text{max}}}}
\newcommand{\aavg}{\ensuremath{a_{\text{avg}}}}
\newcommand{\amin}{\ensuremath{a_{\text{min}}}}
\newcommand{\amax}{\ensuremath{a_{\text{max}}}}
\newcommand{\javg}{\ensuremath{j_{\text{avg}}}}
\newcommand{\jmin}{\ensuremath{j_{\text{min}}}}
\newcommand{\jmax}{\ensuremath{j_{\text{max}}}}

\newcommand{\cdavg}{\ensuremath{\textrm{CD}_{\text{avg}}}}
\newcommand{\cdmax}{\ensuremath{\textrm{CD}_{\text{max}}}}
\newcommand{\svr}{\ensuremath{\textrm{SVR}}}
\newcommand{\ci}{\ensuremath{\textrm{CI}}}
\newcommand{\ssi}{\ensuremath{\textrm{SSI}}}

\newcommand{\ttc}{\ensuremath{\textrm{TTC}_{\text{min}}}}
\newcommand{\ppd}{\ensuremath{\textrm{PPD}}}

\newcommand{\pdce}{\ensuremath{\textrm{pDCE}}}
\newcommand{\intensity}{\ensuremath{\textrm{I}}}
\newcommand{\cp}[1]{\ensuremath{\textrm{CP}_{\text{#1}}}}
\newcommand{\cc}[1]{\ensuremath{\textrm{CC}_{\text{#1}}}}
\newcommand{\resp}[1]{\ensuremath{\textrm{R}_{\text{#1}}}}

\title{\LARGE \bf Spotting the Unfriendly Robot -- \\ Towards better Metrics for Interactions
}

\author{Raphael Wenzel$^{*}$, Malte Probst$^{*}$
\thanks{$^{*}$Honda Research Institute Europe GmbH, Carl-Legien-Str. 30, 63073 Offenbach, Germany, Email:{\tt\footnotesize \{firstname.lastname\}@honda-ri.de}}%
}

\begin{document}

\maketitle
\thispagestyle{empty}
\pagestyle{empty}

\begin{abstract}
Establishing standardized metrics for Social Robot Navigation (SRN) algorithms for assessing the quality and social compliance of robot behavior around humans is essential for SRN research. 
Currently, commonly used evaluation metrics lack the ability to quantify how cooperative an agent behaves in interaction with humans. Concretely, in a simple frontal approach scenario, no metric specifically captures if both agents cooperate or if one agent stays on collision course and the other agent is forced to evade. To address this limitation, we propose two new metrics, a conflict intensity metric and the responsibility metric. Together, these metrics are capable of evaluating the quality of human-robot interactions by showing how much a given algorithm has contributed to reducing a conflict and which agent actually took responsibility of the resolution. This work aims to contribute to the development of a comprehensive and standardized evaluation methodology for SRN, ultimately enhancing the safety, efficiency, and social acceptance of robots in human-centric environments.

\end{abstract}

\section{PROBLEM STATEMENT}\label{sec_prob}\noindent
Research on Social Robot Navigation (SRN) is advancing rapidly, leading to the development of new algorithms that enable intelligent, foresighted navigation around humans, even in dense crowds \cite{francis_principles_2025, gao_evaluation_2022, singamaneni_survey_2024}. However, in order to improve and compare SRN algorithms, clear and agreed upon evaluation protocols and metrics are required. To address this need, recent efforts have been made to establish common ground in the scientific community. Francis et al. \cite{francis_principles_2025} provide a broad survey on the current state of Social Robot Navigation, including evaluation. They surveyed the most prevalent simulators and datasets used for evaluation, along with commonly used metrics and a taxonomy for these metrics. Wang et al. \cite{wang_metrics_2022} propose a condensed set of metrics which evaluate the comfort, naturalness, and sociability along with an evaluation protocol. Gao et al. \cite{gao_evaluation_2022} compile a large number of metrics for navigation performance, human discomfort and sociability. 

All these studies report a lack of agreed-upon benchmarks and evaluation criteria. This holds especially true for the evaluation of one of the core aspects of Social Robot Navigation, the interaction between agents. In this sense, SRN can be seen as a sequence of resolving interactions between various agents until the robot reaches its goal. In each interaction, both parties have to resolve a potential conflict \cite{mirsky_conflict_2024}. Improving the interactions between robots and humans contributes to various aspects of SRN, such as Safety, Social Compliance, or human discomfort. The ability to measure how cooperative an agent behaves around humans is crucial to the development of SRN algorithms, allowing for direct comparison and quantifiable improvement.

This is aggravated by the fact that humans are very adaptive. Humans will adapt quickly to an unresponsive or uncooperative robot by efficiently taking responsibility for the conflict resolution. Even unsophisticated navigation will, most likely, not lead to a significant number of collisions or similar commonly used success metrics for SRN. As a result, most metrics will not capture the difference in social compliance between good or bad robotic behavior. Specifically, when a human and a robot are evading each other, they do not capture the amount to which an agent contributes to the resolution of the conflict. However, socially inept navigation will create inefficiency and annoyance to the humans around the robot, especially if the penetration rate goes up.

We determine that most metrics do not capture complex aspects of behavior quality such as cooperativeness or social compliance. Furthermore, even if those metrics indicate that an observed interaction was resolved proficiently, most metrics are symmetrical by design, i.e., they do not indicate which agent was responsible for the resolution.

To spark discussion on this topic, we show in Section \ref{sec_exp} using a simple experimental setup with highly adaptive agents, that the most commonly used metrics do not capture the social quality of an algorithm's behavior. In Section \ref{sec_adv}, we propose two novel metrics: \emph{Conflict Intensity} and \emph{Responsibility}. We assess their insights into the experiments and discuss potential for further research on advanced metrics to evaluate social compliance.

\section{Existing social compliance metrics}\noindent
In order to evaluate commonly used metrics to assess the performance of SRN algorithms, we use a battery of well-established task-wise metrics. Using the taxonomy established by \cite{francis_principles_2025}, we focus on Social, Hand-Crafted (Algorithmic) and Task-Wise metrics to evaluate quality of behavior and social compliance. Within this category, we distinguish between kinematic, distance-based, and prediction-based. 

Kinematic metrics assess the task performance using the robot's motion data. Commonly used are measures which represent the minimum, average or maximum of the robot's velocity $\vmin$, $\vavg$, $\vmax$, acceleration $\amin$, $\aavg$, $\amax$ and jerk $\jmin$, $\javg$, $\jmax$ \cite{francis_principles_2025}. These measures serve as surrogates for social compliance, based on the assumption that higher social compliance enables the robot to maintain higher speeds or smoother motion profiles.

Metrics based on the robot's distance to pedestrians in each time step capture that close proximity to the robot is inherently dangerous. Francis et al. \cite{francis_principles_2025} collected several quality and social metrics that fall into this category. Measures based on average or maximum \textit{Clearing Distance}, $\cdavg$, $\cdmax$ directly represent the minimum distance between the robot and an object it encounters. Note that the average/maximum refers to the statistics over multiple encounters with several objects. Other measures, like \textit{Space Compliance/Violation Rate} \svr, evaluate the duration (or rate) for which two agents enter each other's designated space. In this context, we consider the space in question to be the Personal Space, according to the Proxemics model \cite{hall_hidden_1966}. Truong and Ngo \cite{truong_dynamic_2016} proposed the \textit{Collision Index} $\ci$ (also referred to as \textit{Social Individual Index} $\ssi$), which is a distance-based metric parameterized by a standard deviation of the pedestrian's personal space.

Some metrics utilize the measured velocities of the robot and other agents, applying constant-velocity predictions to capture the criticality of observed behavior.
The \textit{Minimum Time-To-Collision} \ttc \cite{lee_visual_1977, francis_principles_2025} captures the criticality of a collision if both agents continue to move with their current velocity vector. The \textit{Projected Path Duration} \ppd \cite{jin_mapless_2020, wang_metrics_2022} evaluates the duration, for which the social safety zones of two agents overlap. Each safety zone is represented by a rectangular area in front of the agent defined by the width of the agent and a velocity-proportional length of the safety zone. 

All these metrics aim to capture various aspects of safety, perceived or factual, of behavior computed by SRN algorithms. However, they do not directly capture the key dimensions of social compliance and cooperativity of robot motion planning as the following simulative experiments show. To the best of our knowledge, no existing metrics capture the responsibility assumed by an agent during an interaction.

\section{Experiments}\label{sec_exp}\noindent
In a simple experiment setup, we place two agents 20 meters apart, with their goals set to the other agent's starting position. The objective is to pass each other to reach their goals. Both agents can be either "compliant", i.e., engaging in the interaction by avoiding a collision and accepting a longer path option to reach the goal or "not compliant", by ignoring the other agent and "blindly" showing a constant-velocity behavior. We consider all four permutations of these behavior modes. In the first scenario, both agents follow a direct collision course, passing through each other and reaching their goal on the direct path. While unrealistic, this simulative scenario exemplifies a worst case of bad behavior planning of both agents with the corresponding outcome. In the second scenario, the ego agent (the robot) is non-compliant and proceeds to "barrel through" the other agent to reach its goal. This can be considered bad behavior planning on the side of the robot, as it shows no social compliance whatsoever. In the third scenario, the roles are reversed and the robot is compliant. In Scenarios 2 and 3, no collision occurs due to the high adaptivity and social compliance of one of the agents. In Scenario 4, both agents are compliant, resulting in an efficient resolution of the scenario.

In this experiment, the compliant agents demonstrate foresighted behavior by avoiding each other at an early stage. To achieve this, we parameterize the Social Forces algorithm \cite{moussaid_experimental_2009} with highly cooperative parameters, specifically: $A$~=~5.1, $\lambda~=~3.0$, $\gamma$~=~0.35, $n$~=~1 and $n'$~=~3.0). The results of these simulations, evaluated with the commonly used social evaluation metrics described in Section \ref{sec_prob}, are shown in Table \ref{table_exp1}.

\begin{table*}[ht]
	\label{table_exp1} 
	\caption{Commonly used metrics for evaluating SRN performance}
	\begin{tabular}{@{}crrrrrrrrrrr|rr@{}}
		\toprule
		Scenario & \thead{\scriptsize Resulting \\ \scriptsize paths} &   \thead{\scriptsize (1.)\\\scriptsize Avg.\\ \scriptsize vel} &   \thead{\scriptsize (2.)\\ \scriptsize Avg.\\ \scriptsize accel} &  \thead{\scriptsize (3.)\\ \scriptsize Max. \\ \scriptsize accel} &   \thead{\scriptsize (4.)\\ \scriptsize Avg. \\ \scriptsize jerk} &   \thead{\scriptsize (5.)\\ \scriptsize Max. \\ \scriptsize jerk} &   \thead{\scriptsize (6.) \scriptsize Avg. \\ \scriptsize Clearing \\  \scriptsize Distance} &   \thead{\scriptsize (7.)  Space \\ \scriptsize Violation\\ \scriptsize  Rate} & \thead{\scriptsize (8.)\\ \scriptsize Collision \\ \scriptsize Index} & \thead{\scriptsize (9.)\\\scriptsize Min. \\ \scriptsize TTC } &   \thead{\scriptsize (10.) Proj.\\ \scriptsize Path \\ \scriptsize Duration} &   \thead{\scriptsize \bf Conflict \\ \scriptsize \bf Intensity} &   \thead{\scriptsize \bf Responsibility \\ \scriptsize \bf (R/H)} \\ \midrule
		\makecell{S1: Nobody \\compliant}    & 	\raisebox{-0.3cm}{\resizebox{1.5cm}{!}{\includegraphics{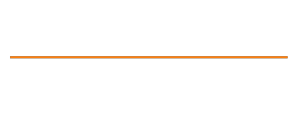}}} & 1.0 & 0.0 & 0.0 & 0.0 & 0.0 & 0.0 & 0.12 & 1.0 & 0.0 & 3.0 & 10.26 & (0,  0) \\ 
		
		\makecell{S2: Robot \\not compliant}    & 	\raisebox{-0.3cm}{\resizebox{1.5cm}{!}{\includegraphics{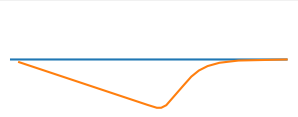}}} & 1.0 & 0.0 & 0.0 & 0.0 & 0.0 & 1.18 & 0.11 & 0.03 & 3.63 & 3.0 & 4.91 & (0,  1) \\
		
		\makecell{S3: Human \\not compliant}    & 		\raisebox{-0.3cm}{\resizebox{1.5cm}{!}{\includegraphics{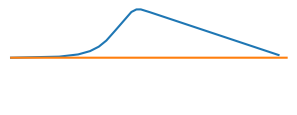}}} & 0.98 & 0.0 & 0.14 & 0.0 & 0.88 & 1.18 & 0.11 & 0.03 & 3.63 & 3.0 & 4.91 & (1,  0)\\
		
		\makecell{S4: Both \\ compliant} & 	\raisebox{-0.3cm}{\resizebox{1.5cm}{!}{\includegraphics{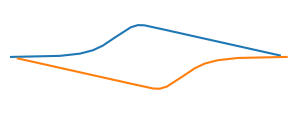}}} & 0.99 & 0.0 & 0.05 & 0.0 & 0.89 & 1.54 & 0.10 & 0.0 & 4.79 & 0.0 & 3.89 & (0.5,  0.5)\\\\ \bottomrule
	\end{tabular}
\end{table*}

\subsection{Finding 1: Evaluating Social Compliance}\noindent
The experiment reveals that none of the metrics reliably distinguish between Scenario 2 (where the robot behaves uncooperatively) and Scenario 4 (where both agents behave cooperatively). The kinematic metrics (1-5) fail to capture the criticality of the robot's behavior because the other agent completely assumes the responsibility of resolving the conflict. Distance-based measures (6-8), such Space Violation Rate $\svr$ and Collision Index $\ci$, exhibit a slight decrease when both agents are cooperative. Consistently, the Clearing Distance $\cdavg$ shows a small increase. However, based on their respective criticality thresholds, none of these metrics would have flagged the robot's behavior as critical. Prediction-based metrics (9-10), such as $\ttc$ and $\ppd$, exhibit the most significant differences, although for incorrect reasons. Although $\ttc$ indicates a difference in criticality, both values are substantially higher than what is normally considered critical. Moreover, the $\ttc$ is only applicable when a collision is predicted, which is too restrictive for movement in 2D space. The $\ppd$ does not detect any criticality when both agents are compliant. In conclusion, none of the commonly employed metrics adequately capture the robot's ability (or inability) to resolve interactions.

\subsection{Finding 2: Evaluating Interaction Responsibility}\noindent
A comparison of the results from Scenario 2 (where the robot behaves uncooperatively) and Scenario 3 (where the human behaves uncooperatively) reveals that none of the evaluated metrics provide a clear indication of which interaction partner took responsibility. The differences in kinematic metrics (1-5) are misleading, showing the uncompliant robot (S2) as moving more smoothly, and will suffer in real-world scenarios that are less sterile. All other metrics (6-10) are, in fact, identical for Scenario 2 and 3. To address this limitation, we introduce two novel metrics in the next section.

\section{PROPOSED METRICS}\label{sec_adv}\noindent

\subsection{Conflict Potential}\noindent
We begin by assessing the conflict potential in a scenario involving two agents: the robot (referred to as "\textit{ego}") and another agent (referred to as "\textit{other}"). Later, we will extend this theory to evaluate entire interactions and explore its application to multiple agents.

Based on \cite{eggert_predictive_2014}, which introduces the concept of Distance at Closest Encounter ($\textrm{DCE}$), we calculate the predicted Distance at Closest Encounter $\pdce$ as a starting point. The $\pdce$ assesses the minimum distance between the two agents when their movements are predicted based on a constant velocity assumption, as can be seen in Figure \ref{fig_dce}. This metric indicates the proximity to a potential collision if both agents continue along their current trajectories. Geometrically, this equates to calculating the perpendicular distance, calculated using the following equation:

\begin{figure}[t]
	\centering
	\includegraphics[width=0.85\columnwidth]{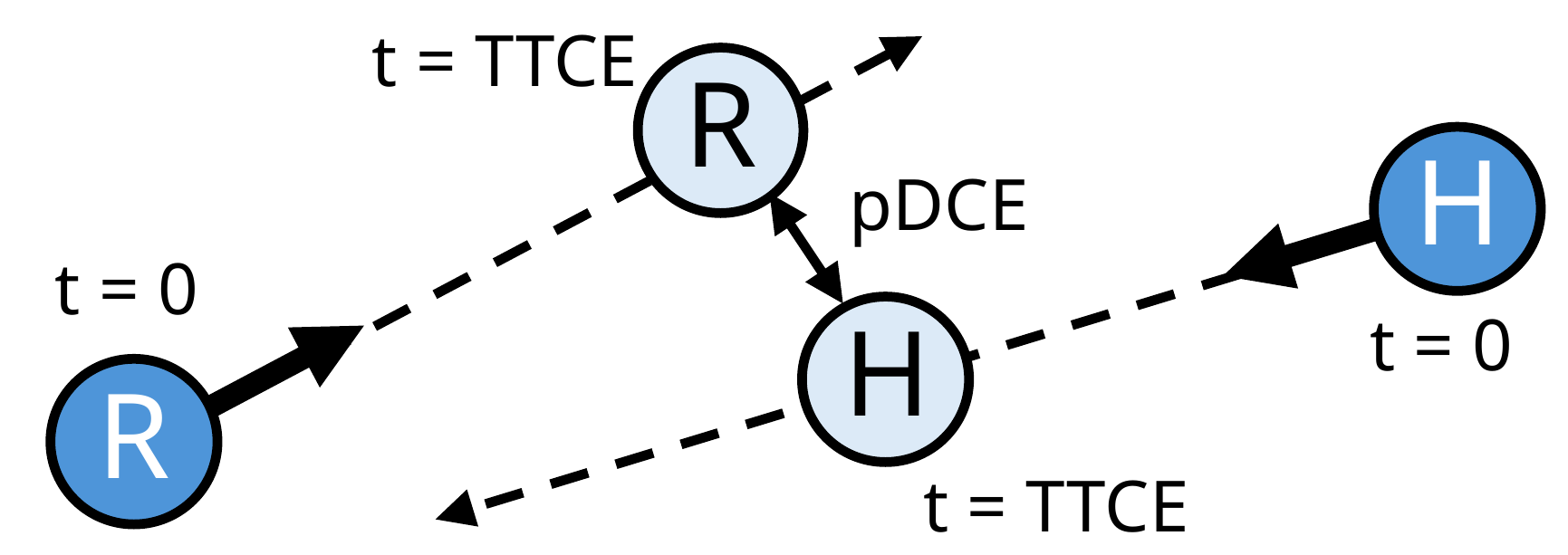}
	\caption{Construction of the predicted Distance at Closest Encounter ($\pdce$). Based on their relative position and velocity, the time to closest encounter (TTCE) can be computed. The distance at that time is the $\pdce$.}

	\label{fig_dce}
\end{figure}

\begin{equation}
	\pdce = \frac{| \mathbf{r} \times \mathbf{v} |}{|\mathbf{v}|} 
\end{equation}

Here, $\mathbf{r}$ is the relative position vector and $\mathbf{v}$ the relative velocity vector of both agents. Based on the $\pdce$, the \textit{Conflict Potential} $\cp{}$ of a situation is given as:

\begin{equation}
	\cp =\max(0, 1 - \frac{\pdce}{s_{\text{ego}} + s_{\text{other}}}) 
\end{equation}

Here, $s_{\text{ego}}$ and $s_{\text{other}}$ represent the radii of both agents. The conflict potential $\cp{}$ indicates the extent of overlap between the two agents at the point of closest encounter. As a result, the conflict potential C is at its max (i.e., equal to 1) in the event of a head-on collision and minimal (i.e., equal to 0) in the case of a near miss. Any motion that alters the agents' trajectory away from a direct collision course will decrease the conflict potential. Therefore, our first proposed metric, the conflict \textit{Intensity} I, is defined as:

\begin{equation}
	I = \int \cp(t) \,dt 
\end{equation}

\subsection{Responsibility}\noindent
We define the \textit{Responsibility} $\resp{}$ in an interaction as the extent to which an agent reduces the conflict potential. To determine an agent's responsibility in conflict resolution, we examine how each agent's behavior $\mathcal{B}$ contributes to reducing the conflict potential $\cp{}$. To approximate this, we calculate the Conflict Contribution $\cc{}$, caused by the change in behavior $d \mathcal{B}$ of an agent in the last time-step, given by

\begin{equation}
	 \cc{agent} = \frac{d}{d \mathcal{B}} \cp{} \approx \cp{} - \cp{no change, agent} \quad,
\end{equation}

where $\cp{no change, agent}$ is calculated based on the agent's velocity vector $v^{t-1}$ from the previous time step. By integrating the Conflict Contribution $\cc{}$ over time for each agent, we obtain a measure of the agent's contribution to conflict resolution, as expressed by:

\begin{equation}
	\label{eq_resp}
	 R = \frac{1}{\cp{0}}\int \cc{}(t) \,dt
\end{equation}

Here, $\cp{0}$ is the conflict potential at the start of the interaction which must be reduced to resolve the conflict. Both the conflict Intensity $I$ and the agent's Responsibility $R$ are task-wise metrics, providing a single scalar value that characterizes the entire observed interaction. In contrast, the intermediate quantities $\cp{}$ (conflict potential) and $\cc{}$ (conflict contribution) are step-wise metrics, whose evolution over time provides insights into the progression of the interaction between the two agents.

\section{Results for proposed metrics}\noindent
For Scenario 1, a head-on collision scenario, the \textit{Conflict Intensity} for both agents is $I=$~10.255. If either of the agents takes steps towards resolving the interaction, the conflict intensity decreases to $I=$ 4.910, and decreases to $\intensity=$~3.889 if both agents cooperate. The relativity between these values appears to be consistent, reflecting the fact that the conflict is present in all cases, but the resolutions differ.
We argue that this is beneficial for measuring compliance in social robot navigation, as it reflects the persistence of underlying conflicts despite mitigation by one of the agents. 
The reduction of Intensity from Scenarios 2/3 to 4 is relatively small. This is plausible, as one agent would have resolved the conflict independently. However, when both agents share the burden, the intensity is reduced further, albeit only slightly.

In contrast, other metrics seem to be less consistent across scenarios. $\cdavg$ and $\ttc$ show a drastic reduction in criticality between Scenario 1 vs. Scenario 2/3. Metrics like $\ppd$ and $\ci$ decrease sharply when both agents resolve the conflict, despite all scenarios beginning with the same initial conflict. The $\svr$ shows only a very small relative change across all 4 Scenarios. This makes it harder to compare different scenarios or algorithms.

The Responsibility metric shows a clear difference between the four scenarios: In the first scenario, neither agent takes responsibility for avoiding the collision, resulting in a collision. The Responsibility for both agents evaluates to $\resp{R}$ = $\resp{H}$ = 0.0. In Scenarios 2 and 3, the agent that takes responsibility earns the full share of the Responsibility metric. Specifically, $\resp{H}$ = 1.0 (in Scenario 3) and $\resp{R}$ = 1.0 (in Scenario 2), respectively. In the fourth scenario, where both agents share the responsibility equally, the Responsibility metric clearly indicates the cooperation: $\resp{R}$ = $\resp{H}$~=~0.5. The equal 50\% share of responsibility between the two agents is a logical consequence of the symmetrical initial conditions and the identical behavior planners used for both agents.

\begin{figure}[t]
	\centering
	\includegraphics[width=0.95\linewidth]{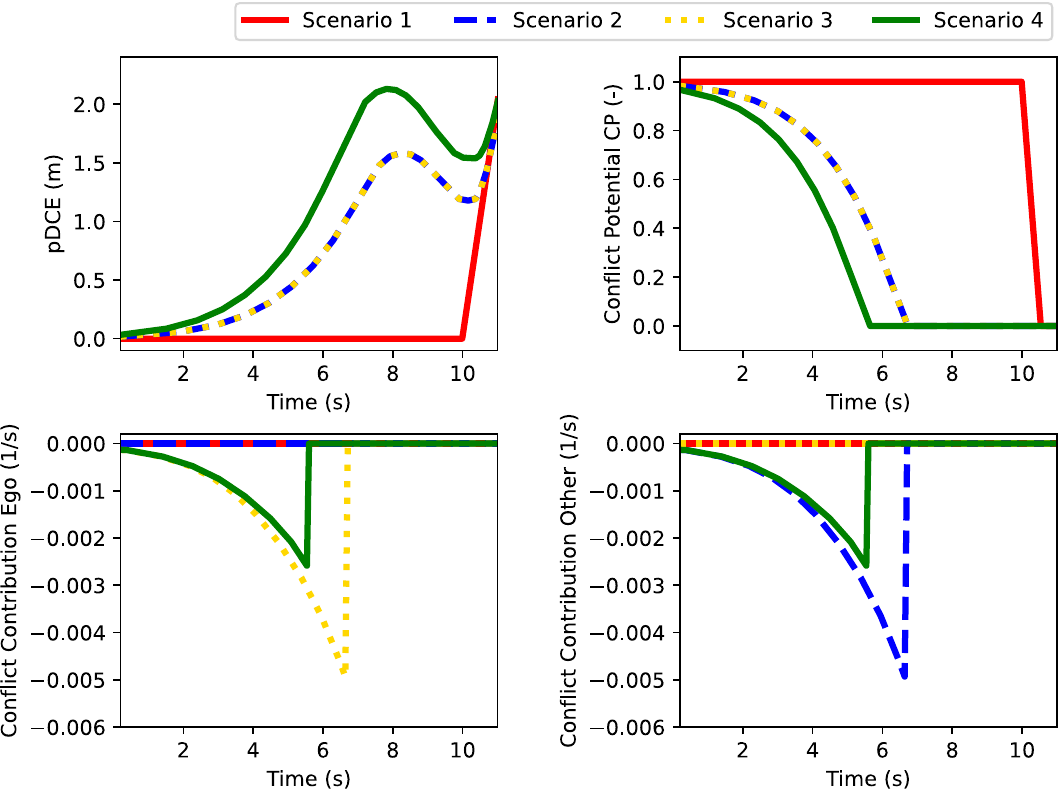}
	\caption{Progression in all 4 Scenarios of various step-wise metrics used in this work to derive the Intensity and Responsibility metrics.}

	\label{fig_progression}
	\vspace*{-5mm}
\end{figure}

The proposed step-wise quantities Conflict Potential ($\cp{}$) and Conflict Contribution ($\cc{}$) also provide insights into the conflict resolution in these exemplary scenarios. Figure \ref{fig_progression} provides the progression of CP and CC values in all 4 Scenarios. The top right plot shows that, while all scenarios begin with the same Conflict Potential, the rate of conflict resolution varies significantly. In Scenarios 2 and 3, the conflict potential decreases at the same rate. However, in Scenario 4, both agents cooperate to resolve the interaction, resulting in a steeper decline in Conflict Potential and an earlier resolution. Since the Intensity of the conflict is equivalent to the area under the Conflict Potential curves, this relationship is also reflected in the Intensity metric. Similarly, the Conflict Contributions (lower left: "ego, lower right: "other") is smaller in Scenario 4, where both agents cooperate. Consequently, the responsibility assigned to both agents is smaller, as it is normalized by the initial overall intensity of the interaction.

\section{Discussion}
\noindent
To effectively use the proposed metrics as social quality metrics for SRN, they must be applied not only to pairwise interactions but also consider the presence of all other pedestrians in the scene. To achieve this, average values of Intensity and Responsibility can be used to assess how a given algorithm's behavior affects conflict intensity, either by reducing or increasing it. Similarly, the average Responsibility can serve as an indicator of the robot's contribution to conflict resolution.

Importantly, both metrics can be parameterized with the agent radii to focus on social compliance regarding collision or invasion of personal space. This enables a more nuanced assessment of agent behavior, based on various environmental conditions, such as crowd density.

A key insight from the construction of collision intensity is that it is an integral over the collision potential. This indicates that early conflict resolution significantly reduces the intensity of a conflict compared to a late resolution. This aligns with intuition, where a foresighted action reduces the overall criticality compared to a late reaction, even though the final outcome is the same. Moreover, the responsibility metric would also attribute a significant share to an agent executing an early action.

The proposed metrics are designed to be generalizable. The use of $\pdce$ as a base for constructing the metrics makes the metrics meaningful for all motion in 2D space. Additionally, the metrics can capture various types of behavior $\mathcal{B}$. In an additional experiment, we observed that two Social Force agents with the same parameters crossing at a 90° angle showed different reactions to mitigate the collision. One agent slowed down, resolving the interaction via a change in speed. The other agent evaded by veering away, resolving the interaction by changing direction. In this case, the responsibility according to Eq. \ref{eq_resp} showed an almost equal share of responsibility.

\section{Conclusion}\noindent
This paper addresses the need for advanced evaluation metrics that capture social compliance and cooperativity in Social Robot Navigation. Humans are highly adaptive around robots, which is one of the main reasons why commonly used metrics are insufficient. Through simulative experiments, we demonstrate that typical metrics struggle to distinguish between simplistic and socially compliant navigation algorithms when interacting with cooperative partners. We identify a need for additional metrics that capture the reduction of conflict intensity and the allocation of responsibility between agents. We propose two metrics, the \emph{Conflict Intensity} and the \emph{Responsibility} for an agent's observed behavior. We show that these metrics effectively capture the observed effects in our simulative experiments. Our goal is to initiate a discussion on the development of suitable metrics for evaluating the performance of Social Robot Navigation algorithms in human-robot interactions, as well as to explore more options for capturing the interaction quality, e.g., with information-theoretic metrics on cooperation \cite{wollstadt_quantifying_2022,wollstadt_quantifying_2023}.

\bibliographystyle{IEEEtran}
\bibliography{root}

\end{document}